\documentclass[conference]{IEEEtran}
\usepackage{ifpdf}
\usepackage{cite}
\ifCLASSINFOpdf
   \usepackage[pdftex]{graphicx}
\else
   \usepackage[dvips]{graphicx}
\fi
\usepackage{amsmath}
\usepackage{algorithmic}
\usepackage{array}
\ifCLASSOPTIONcompsoc
  \usepackage[caption=false,font=normalsize,labelfont=sf,textfont=sf]{subfig}
\else
  \usepackage[caption=false,font=footnotesize]{subfig}
\fi
\usepackage{url}
\usepackage{multirow}
\usepackage{multicol}
\usepackage{booktabs} 
\usepackage{color}

\begin{document}
%
% paper title
% Titles are generally capitalized except for words such as a, an, and, as,
% at, but, by, for, in, nor, of, on, or, the, to and up, which are usually
% not capitalized unless they are the first or last word of the title.
% Linebreaks \\ can be used within to get better formatting as desired.
% Do not put math or special symbols in the title.
\title{Correlation Heuristics for Constraint Programming}

% author names and affiliations
% use a multiple column layout for up to three different
% affiliations

\author{
\IEEEauthorblockN{Ruiwei Wang}
\IEEEauthorblockA{School of Computing \\ National University of Singapore \\
Email: wangruiw@comp.nus.edu.sg}
\and
\IEEEauthorblockN{Wei Xia}
\IEEEauthorblockA{School of Computing \\ National University of Singapore \\
Email: xiawei@comp.nus.edu.sg}
\and
\IEEEauthorblockN{Roland H. C. Yap}
\IEEEauthorblockA{School of Computing\\
National University of Singapore \\
Email: ryap@comp.nus.edu.sg}
 }

% conference papers do not typically use \thanks and this command
% is locked out in conference mode. If really needed, such as for
% the acknowledgment of grants, issue a \IEEEoverridecommandlockouts
% after \documentclass

% for over three affiliations, or if they all won't fit within the width
% of the page, use this alternative format:
% 
%\author{\IEEEauthorblockN{Michael Shell\IEEEauthorrefmark{1},
%Homer Simpson\IEEEauthorrefmark{2},
%James Kirk\IEEEauthorrefmark{3}, 
%Montgomery Scott\IEEEauthorrefmark{3} and
%Eldon Tyrell\IEEEauthorrefmark{4}}
%\IEEEauthorblockA{\IEEEauthorrefmark{1}School of Electrical and Computer Engineering\\
%Georgia Institute of Technology,
%Atlanta, Georgia 30332--0250\\ Email: see http://www.michaelshell.org/contact.html}
%\IEEEauthorblockA{\IEEEauthorrefmark{2}Twentieth Century Fox, Springfield, USA\\
%Email: homer@thesimpsons.com}
%\IEEEauthorblockA{\IEEEauthorrefmark{3}Starfleet Academy, San Francisco, California 96678-2391\\
%Telephone: (800) 555--1212, Fax: (888) 555--1212}
%\IEEEauthorblockA{\IEEEauthorrefmark{4}Tyrell Inc., 123 Replicant Street, Los Angeles, California 90210--4321}}

% use for special paper notices
%\IEEEspecialpapernotice{(Invited Paper)}

% make the title area
\maketitle

% As a general rule, do not put math, special symbols or citations
% in the abstract
\begin{abstract}
Effective general-purpose search strategies are an important component in Constraint Programming.
We introduce a new idea, namely, using correlations between variables to 
guide search. 
Variable correlations are measured and maintained by using
domain changes during constraint propagation. 
We propose two variable heuristics based on the correlation matrix,
\emph{crbs-sum} and \emph{crbs-max}.
We evaluate our correlation heuristics with well known heuristics, namely, \textit{dom/wdeg}, impact-based search and activity-based search.
Experiments on a large set of benchmarks show that our correlation heuristics
are competitive with the other heuristics, 
and can be the fastest on many series.
\end{abstract}

% no keywords

% For peer review papers, you can put extra information on the cover
% page as needed:
% \ifCLASSOPTIONpeerreview
% \begin{center} \bfseries EDICS Category: 3-BBND \end{center}
% \fi
%
% For peerreview papers, this IEEEtran command inserts a page break and
% creates the second title. It will be ignored for other modes.
\IEEEpeerreviewmaketitle

\section{Introduction}
Backtracking search combined with constraint solving is the
main approach to solve problems in Constraint Programming (CP).
The key to effective search is having a good variable search heuristic to select
a variable to branch as 
the size of the search tree is strongly dependent on the selected variables. 
In CP, many general purpose variable ordering search heuristics 
have been proposed and implemented in many CP solvers, 
such as the conflict-driven heuristic \textit{dom/wdeg} \cite{wdeg}, impact-based search (\textit{IBS}) heuristic \cite{impact}, and activity-based search (\textit{ABS}) heuristic \cite{activity}.
Search heuristics by their nature are not designed to be optimal search strategies but merely good ones.
Thus, our goal in this paper is a new search heuristic which can outperform existing heuristics on some instances across a range of problems.

We propose a new idea which is correlation-based search (\textit{CRBS}), 
the search heuristic employs correlations between variables.
The correlation of a pair of variables ($x_i$, $x_j$) is used to estimate the possibility of a conflict between $x_i$ and $x_j$ during search.
We maintain a matrix corresponding to the paired variable correlation 
during search.
The correlation matrix is turned into a search strategy by using
a function to combine values in the matrix
to estimate whether assigning a value to variable $x_i$ can cause a conflict.
Domain changes during constraint propagation are used
to measure the correlations between variables. 
We present two generic and new correlation-based variable heuristics, 
\emph{crbs-sum} and \emph{crbs-max}.
Our experiments compare the correlation heuristics with 
the well known search heuristics \textit{dom/wdeg}, \textit{ABS} and \textit{IBS} on a large set of benchmarks.
The results show that correlation heuristics are competitive 
with the existing heuristics, and
can also be the fastest on many problem instances from different problem
series.
In particular, \emph{crbs-sum} is shown to be an effective search heuristic.

\section{Related works}
We briefly introduce several well-known general purpose heuristics. 
One of the simplest heuristics is \textit{dom}\cite{dom} 
which follows the fail first principle, selecting the variable with smallest
domain size.
Many general purpose heuristics combine domain size with other information.
For example, the well-known heuristics \textit{dom/deg} \cite{domOverDeg} and \emph{dom/ddeg} \cite{ddeg} combine domain sizes with variable degrees, which can be better than \textit{dom}.
The conflict-driven heuristic \textit{dom/wdeg} \cite{wdeg} 
associates a weight with each constraint to record conflicts during search.
The weight of constraint $c$ is increased when the constraint solver finds 
$c$ to be inconsistent.
The \textit{dom/wdeg} heuristic selects the next variable based on weight degrees and domain sizes, where the weight degree of a variable $x$ is 
the sum of the weights of the constraints involving $x$ and at
least another uninstantiated variable.
Some variants of \textit{dom/wdeg} exploit different information to update the weight of constraints such as the explanation-based weight \cite{ewdeg} and constraint tightness weight \cite{wtdeg}.

The impact-based search (\textit{IBS}) heuristic \cite{impact} is motivated 
by the pseudo-costs used in mixed-integer programming.
It uses impact to measure the importance of a variable to the rate of 
search space reduction.  
A variant of \textit{IBS} incorporates variances in reduction \cite{impactv}.
Counting-based search \cite{counting} exploits solution counting information to guide search. 
The activity-based search (\textit{ABS}) heuristic \cite{activity} 
combines domain sizes with activity for variables where
activity is a measure of how often a variable is reduced during search.
We remark that it is different from the SAT activity heuristic 
\textit{VSIDS} \cite{vsids} which also records some conflict information 
during search.

\section{Background}
A constraint satisfaction problem (CSP) instance is a triplet $(C,X,D)$, 
where $C=\{c_1, c_2, ... c_e\}$ is a set of $e$ constraints, $X=\{x_1, 
x_2, ...x_n\}$ is a set of $n$ variables and $D=\{D(x_1), D(x_2), ... D(x_n)\}$ 
is the corresponding domains for the variables.
$D(x_i)$ is the initial domain of variable $x_i$, and $dom(x_i)\subseteq 
D(x_i)$ is the current domain of $x_i$ during search.
Every constraint $c$ consists of a constraint scope $scp(c)$ and a relation $R(c)$, where $scp(c)\subseteq X$ and $R(c)\subseteq \prod\limits_{x_i\in scp(c)}D(x_i)$.
A solution of a CSP instance is the set of assignments $\{(x_1,a_1), ... 
(x_n,a_n)\}$ which satisfies all constraints in $C$, where $a_i\in 
D(x_i)$.
During backtrack search, the search heuristic selects a variable to 
instantiate at each search node. The variables which have been 
instantiated during a path in the search tree
are defined as \emph{past} variables while the variables 
which have not been instantiated are \emph{future} variables. 

\section{Correlation-based search}
Typically the goal of a variable heuristic is to choose variables
which can cause backtracking to occur earlier in the search.
This suggests to choose variables which can
lead to conflicts earlier in the search.
In this paper, we propose correlation-based heuristics to 
achieve this objective.
For each pair of variables $(x_i,x_j)$, we define a value $a_{i,j}$, called the 
\emph{correlation of $(x_i,x_j)$}, as a measure of the possibility of 
having a conflict between $x_i$ and $x_j$.
During search, a \emph{correlation matrix} representing all variable pairs
$a_{i,j}$ is maintained,
where each value in the matrix represents the correlation of a 
pair of variables.
A special case is $a_{i,i}$ which estimates the degree of conflict
when choosing variable $x_i$.
We propose two functions which use the
correlation matrix to estimate the degree of conflict from assigning
the variable.
Then the heuristic will choose the variable which is estimated to
cause more conflicts.

\subsection{Updating the correlation matrix}
We maintain the correlation matrix by using domain changes during constraint propagation. 
Some search heuristics have used the information about domain changes to guide search,
such as activity-based search (\textit{ABS}) \cite{activity}. 
The idea of the \textit{ABS} heuristic is to select the variable which is the most often updated. 
It maintains an array $A$ during search to record the activities of variables. 
After constraint propagation, if the domain of variable $x_i$ is updated, then $A(x_i)$ is increased by 1, otherwise decreased by multiplying with $\gamma$
where $0 \le \gamma \le 1$. 
Then the heuristic selects the variable with maximal $A(x_i)/dom(x_i)$.

We use a similar approach.
We assume that the more frequent $dom(x_j)$ is updated after assigning $x_i$, 
the more likely a conflict between $x_i$ and $x_j$ can happen.
As such, the correlations between variables are updated based on 
domain changes.
After constraint propagation due to variable $x_i$ being assigned, 
the remaining variables can be split into two subsets, $U$ and $N$: 
\begin{displaymath}
\begin{aligned}
U=\{\forall x_j\in X'~|~dom'(x_j)\neq dom(x_j)\}\\
N=\{\forall x_j\in X'~|~dom'(x_j)= dom(x_j)\}\\
\end{aligned}
\end{displaymath}
where $X'=X\setminus\{x_i\}$ and $dom'(x_j)$ is the new domain of $x_j$ after constraint propagation. The $U$ variables are those whose domains are updated, 
while the $N$ variables are those whose domains are unchanged.
If no conflict occurs, then the correlations are updated 
as follows:
\begin{equation}
\left\{
\begin{aligned}
a_{i,j}=\check{a}_{i,j}+1,a_{j,i}=\check{a}_{j,i}+1\hspace{0.68cm}\forall x_j\in U\\
a_{i,j}=\check{a}_{i,j}-1,a_{j,i}=\check{a}_{j,i}-1~~~~~\forall x_j\in N\\
a_{i,i}=\check{a}_{i,i}-1\hspace{4.16cm}\\
\end{aligned}
\right.
\end{equation}
where $\check{a}_{i,j}$ is the old correlation value before the update.
If $dom'(x_j)$ is changed after assigning $x_i$, then the correlations $a_{i,j}$ and $a_{j,i}$ are increased by one.
Otherwise, $a_{i,j}$ and $a_{j,i}$ are decreased by one.
In addition, we decrease the correlation $a_{i,i}$, 
this is to make $a_{i,i}$ small if no conflicts happen after 
assigning $x_{i}$ repeatedly.

Otherwise, if a conflict appears in the constraint propagation after assigning $x_i$, the correlations of all variables are increased as follows:
\begin{equation}
\left\{
\begin{aligned}
a_{i,j}=\check{a}_{i,j}+1,a_{j,i}=\check{a}_{j,i}+1~~~~~\forall x_j\in X'\\
a_{i,i}=\check{a}_{i,i}+2\hspace{4.26cm}\\
\end{aligned}
\right.
\end{equation}
We increase the correlation $a_{i,i}$ by 2 because the assignment of $x_{i}$ 
causes a conflict.
In addition, $a_{i,j}$ and $a_{j,i}$ are updated in the same way as before.
We see that this definition leads to the correlation matrix being symmetric. 

\subsection{Selecting variables using the correlation matrix}
We propose two ways of using the correlation matrix with combining
functions based on the matrix and problem variables, namely,
the \textit{crbs-sum} and \textit{crbs-max} functions which estimate
the potential of conflict after assigning variable $x_i$.

The \textit{crbs-sum} function is a linear function of the relevant
entries in the correlation matrix.
First, we define two auxiliary functions,
$P_c(x_i)$ and $F_c(x_i)$ on variable $x_i$:
\begin{displaymath}
\begin{aligned}
P_c(x_i)=\sum\limits_{x_j\in P}a_{i,j}&   ~~~~~~~     &F_c(x_i)=\sum\limits_{x_j\in F}a_{i,j}
\end{aligned}
\end{displaymath}
where $P$ is a set of past variables and $F$ is a set of future variables. 
The variable to be considered, $x_i$ is part of the set $F$ of future
variables.
The idea is that
$P_c(x_i)$ (past correlation) is the sum of correlations of past variables
with respect to $x_i$,
and $F_c(x_i)$ (future correlation) is similar but for
the future variables.
The \textit{crbs-sum} function for variable $x_i$ is defined as:
\begin{equation}
	\mbox{\it crbs-sum}(x_i)=P_c(x_i)~+~\theta\times F_c(x_i)
\label{cor}
\end{equation} 
A parameter $0\le\theta\le1$ is used to control the combination of
the past and future variable correlation.
In particular, future variables are used when $\theta>0$, 
otherwise, we consider only past variables when $\theta=0$.

We propose another simple combining function,
the \textit{crbs-max} function, defined as follows: 
\begin{equation}
	\mbox{\it crbs-max}(x_i)=\max\limits_{x_j\in P}(a_{i,j})
	\label{eq:max}
\end{equation}  
The idea for \textit{crbs-max} is to choose a future variable which has 
the largest estimated correlation with the past variables.
We also experimented with a variant of the max function on all variables
(past and future),
i.e. $\max\limits_{x_j\in X}(a_{i,j})$.
Initial experiments found Equation \ref{eq:max} to give better results.
In the rest of the paper, we use the max function as defined in
Equation \ref{eq:max}.

The variable chosen by either the \textit{crbs-sum} or \textit{crbs-max} heuristic
is the variable $x_i$ which maximizes $f(x_i)/dom(x_i)$
where $f$ is either the
\textit{crbs-sum} or \textit{crbs-max} function.

\begin{table*}
\renewcommand{\arraystretch}{0.9}
\small
\begin{center}
\tabcolsep=0.7mm
\vspace{0.4in}
\centering
\begin{tabular}{l|l|rrrrr|rrrrr}
\toprule [1 pt] 
\multirow{2}{*}{series}&\multirow{2}{*}{}&\multicolumn{5}{c|}{mean time (s)}&\multicolumn{5}{c}{\textit{nodes}}\\
\cline{3-12}
&&\textit{dom/wdeg}&\textit{ABS}&\textit{IBS}&\textit{crbs-sum}&\textit{crbs-max}&\textit{dom/wdeg}&\textit{ABS}&\textit{IBS}&\textit{crbs-sum}&\textit{crbs-max}\\
\toprule [1 pt] 
\toprule [1 pt] 
Ortholatin&total (4)&107.84 &1 TO&1 TO&1 TO&\textbf{31.75} &515K&-&-&-&\textbf{87K}\\
&solved by all (3)&2.09 &0.49 &9.42 &1.41 &0.70 &11K&868 &52K&6K&1K\\
\toprule [1 pt] 
\toprule [1 pt] 
TSP&total (30)&5.48 &13.12 &12.55 &\textbf{3.96} &7.93 &44K&228K&253K&\textbf{32K}&74K\\
&solved by all (30)&5.48 &13.12 &12.55 &3.96 &7.93 &44K&228K&253K&32K&74K\\
\hline
Latin Square&total (6)&1 TO&1 TO&26.73 &\textbf{0.77} &1.43 &-&-&334K&\textbf{5K}&10K\\
&solved by all (5)&0.27 &0.27 &0.27 &0.28 &0.30 &31 &22 &31 &44 &117 \\
\hline
Dubois&total (11)&4 TO&4 TO&3 TO&\textbf{17.58} &71.14 &-&-&-&\textbf{2M}&3M\\
&solved by all (7)&273.84 &222.59 &128.63 &5.11 &30.72 &57M&30M&38M&\textbf{979K}&1M\\
\hline
Magic Square&total (11)&4 TO&5 TO&4 TO&\textbf{172.16} &1 TO&-&-&-&\textbf{566K}&-\\
&solved by all (6)&36.48 &1.06 &26.94 &0.47 &1.65 &254K&4K&225K&1K&8K\\
\hline
Costas Array&total (9)&1 TO&2 TO&1 TO&\textbf{111.05} &1 TO&-&-&-&\textbf{158K}&-\\
&solved by all (7)&3.30 &3.44 &8.62 &1.54 &5.65 &5K&5K&18K&2K&8K\\
\hline
Social Golfer&total (4)&1 TO&2 TO&1 TO&\textbf{52.16} &3 TO&-&-&-&\textbf{101K}&-\\
&solved by all (0)&- &- &- &- &- &-&-&-&-&-\\
\hline
ii&total (41)&1 TO&3.63 &2 TO&\textbf{3.11} &2 TO&-&2K&-&\textbf{1K}&-\\
&solved by all (38)&10.34 &3.54 &30.49 &1.93 &111.14 &8K&2K&51K&526 &699 \\
\hline
Register&total (8)&2 TO&2 TO&1 TO&\textbf{431.66} &4 TO&-&-&-&\textbf{1M}&-\\
&solved by all (4)&0.77 &0.72 &0.94 &0.68 &0.82 &158 &125 &467 &95 &170 \\
\hline
Quasi Group&total (25)&15.43 &15.51 &48.20 &\textbf{9.36} &17.91 &84K&114K&471K&\textbf{52K}&62K\\
&solved by all (25)&15.43 &15.51 &48.20 &9.36 &17.91 &84K&114K&471K&52K&62K\\
\hline
Super-jobShop&total (22)&2 TO&3 TO&2 TO& \textbf{2 TO} &5 TO&-&-&-&\textbf{-}&-\\
&solved by all (16)&1.36 &1.38 &1.73 &1.40 &1.42 &389 &140 &6K&\textbf{94} &139 \\
\toprule [1 pt] 
\toprule [1 pt] 
Nonogram&total (176)&4 TO&1.63 &\textbf{1.59} &1.61&4.89 &-&564 &181 &\textbf{102} &177 \\
&solved by all (172)&5.96 &1.59 &1.57 &1.59 &4.46 &62K&342 &172 &97 &158 \\
\hline
Cril&total (8)&2 TO&2 TO&\textbf{2.53} &3.36 &23.62 &-&-&\textbf{82K}&120K&1M\\
&solved by all (6)&2.56 &1.72 &3.18 &4.33 &15.46 &310K&94K&110K&160K&2M\\
\hline
Black hole&total (39)&19 TO&27.72 &\textbf{15.92} &34.04 &1 TO&-&3M&\textbf{1M}&4M&-\\
&solved by all (20)&161.23 &0.34 &0.39 &0.34 &0.34 &25M &40 &1613 &30 &29 \\
\hline
Myciel&total (12)&12.36 &8.71 &\textbf{5.53} &6.65 &1 TO&429K&234K&\textbf{134K}&194K&-\\
&solved by all (11)&6.78 &3.11 &2.06 &3.73 &68.66 &392K&150K&87K&170K&2M\\
\hline
Queen Knights&total (11)&2 TO&4 TO&\textbf{78.81} &3 TO&3 TO&-&-&\textbf{3K}&-&-\\
&solved by all (8)&37.77 &181.06 &5.77 &174.54 &234.65 &5K&29K&769 &24K&34K\\
\hline
AllInterval&total (9)&1 TO&1 TO&\textbf{17.86} &71.34 &48.28 &-&-&\textbf{470K}&1M&1M\\
&solved by all (8)&4.00 &30.06 &0.91 &4.61 &1.21 &79K&1M&19K&110K&22K\\
\hline
cc&total (13)&1 TO&1 TO&\textbf{24.77} &1 TO&2 TO&-&-&\textbf{118K}&-&-\\
&solved by all (11)&3.43 &2.70 &6.22 &2.80 &3.07 &2K&1K&3K&1K&2K\\
\hline
Open Shop&total (49)&60.95 &1 TO&\textbf{51.00} &2 TO &5 TO&\textbf{55K}&-&80K&-&-\\
&solved by all (43)&39.83 &79.36 &47.33 &52.54 &79.11 &17K&16K&38K&19K&159K\\
\hline
coloring&total (22)&1.87 &0.65 &\textbf{0.62} &0.99 &7.88 &144K&34K&\textbf{20K}&55K&493K\\
&solved by all (22)&1.87 &0.65 &0.62 &0.99 &7.88 &144K&34K&20K&55K&493K\\
\toprule [1 pt] 
\toprule [1 pt] 
Mug&total (8)&4 TO&\textbf{31.57} &3 TO&173.65 &3 TO&-&\textbf{7M}&-&28M&-\\
&solved by all (4)&0.33 &0.33 &0.33 &0.33 &0.34 &0 &0 &0 &0 &0 \\
\hline
Knights&total (8)&1 TO&\textbf{158.37} &171.04 &172.61 &173.73 &-&\textbf{1K}&\textbf{1K}&\textbf{1K}&\textbf{1K}\\
&solved by all (7)&41.85 &29.40 &28.62 &28.89 &33.52 &1K&934 &934 &934 &934 \\
\hline
Covering Array&total (9)&4 TO&\textbf{2.71} &1 TO&3.77 &3 TO&-&\textbf{3K}&-&4K&-\\
&solved by all (5)&79.40 &0.46 &0.48 &0.45 &9.98 &778K&219 &646 &128 &7K\\
\hline
Insertion&total (21)&1 TO&\textbf{4.12} &5 TO&8.37 &3 TO&-&\textbf{187K}&-&523K&-\\
&solved by all (14)&1.31 &0.86 &8.64 &1.48 &1.58 &3K&832 &61K&823 &5K\\
\toprule [1 pt]  
\toprule [1 pt]  
Radar&total (62)&\textbf{8.77} &23.96 &4 TO&43.09 &38.63 &\textbf{107} &631 &3K&1K&583 \\
&solved by all (58)&4.54 &10.90 &45.95 &24.38 &24.66 &60 &523 &1K&618 &397 \\
\hline
Queen Attack&total (5)&\textbf{2.33} &2 TO&1 TO&23.64 &1 TO&\textbf{41K}&-&-&579K&-\\
&solved by all (3)&0.37 &0.37 &0.48 &0.34 &1.17 &483 &914 &5K&219 &40K\\
\hline
scen11&total (10)&\textbf{45.24} &4 TO&1 TO&60.65 &3 TO&1M&-&-&\textbf{963K}&-\\
&solved by all (6)&0.98 &0.96 &1.42 &1.08 &11.68 &3K&3K&15K&2K&114K\\
\hline
Crossword&total (140)&\textbf{2.08} &7.63 &12 TO &3.77 &2.88 &\textbf{12K}&73K&-&26K&16K\\
&solved by all (128)&1.17 &6.09 &59.54 &2.25 &1.70 &10K&71K&643K&22K&13K\\
\hline
Golomb Ruler&total (25)&\textbf{1 TO}&7 TO&5 TO&6 TO&5 TO&\textbf{-}&-&-&-&-\\
&solved by all (18)&25.89 &70.15 &38.17 &48.51 &23.59 &\textbf{57K}&150K&90K&99K&31K\\
\hline
Schurr Lemma&total (9)&\textbf{52.85} &1 TO&2 TO&81.77 &1 TO&\textbf{144K}&-&-&416K&-\\
&solved by all (7)&18.14 &29.25 &33.54 &16.55 &27.93 &156K&211K&178K&143K&176K\\
\toprule [1 pt] 
\toprule [1 pt] 
\multirow{2}{*}{Total}&total (807)&56 TO& 43 TO&48 TO&\textbf{15 TO}&47 TO&-&-&-&-&-\\
&solved by all (692)&16.08&14.34&25.96&\textbf{9.53}&20.35&1M&364K&549K&\textbf{30K}&108K\\
\toprule [1 pt] 
\end{tabular}
\caption{Mean results of 5 heuristics. For the Super-jobShop series, \textit{crbs-sum} is highlighted as the best it has the smallest total runtime on solving the 20 non-timeout instances compared with \textit{dom/wdeg} and \textit{IBS}.}  \label{exp:table1} 
\end{center}
\end{table*}
\section{Experiments}
We evaluate the correlation-based heuristics, \textit{crbs-sum} and 
\textit{crbs-max},
with well known, successful and commonly used heuristics:
weighted degree (\textit{dom/wdeg}), activity (\textit{ABS}) and impact 
(\textit{IBS}).
Experiments are run on a 3.40 GHz Intel core i7 CPU on Linux.
The existing heuristics are the 
AbsCon\footnote{
We used the AbsCon solver (\url{https://www.cril.univ-artois.fr/~lecoutre/software.html}).
} 
solver implementations of \textit{dom/wdeg}, \textit{ABS} and \textit{IBS}.
For the \textit{ABS} and \textit{IBS} heuristics, we use the default parameter settings in Abscon.

For the crbs-sum heuristic, we use $\theta=0.1$,
chosen as a value for $\theta$ which we found to work well
on many instances
(see Section \ref{sec:parameter}).
The initial values in the correlation matrix of CRBS are set to 0.

All heuristics break ties lexicographically, and use the
lexical value order heuristic.
In all cases, a geometric restart search policy (the initial \textit{cutoff}=10 and $\rho=1.1$) was used, where \textit{cutoff} is the maximum number of failures before restart and $\rho$ controls the growth of the value of \textit{cutoff} after restart.\footnote{The value of \textit{cutoff} is updated using $\textit{cutoff}=\textit{cutoff'}+init\_\textit{cutoff} * {\rho}^k$,  where \textit{cutoff'} is the cutoff of the last restart, $init\_\textit{cutoff}$ is the initial cutoff with value 10, and $k$ is the number of encountered restarts.}
We apply the binary search branch strategy.
The time-out is set to 1200s for all instances.
We have used a large and varied set of well-known CSP benchmarks.\footnote{Benchmarks are from the 2009 CSP competition website: \url{http://www.cril.univ-artois.fr/CSC09/} and the XCSP3.0 website \url{http://www.xcsp.org/}}
In total, there are 807 problem instances which come from the following 30 
series:
\begin{quote}
All Interval Series (AllInterval), Black Hole, Chessboard Coloration (cc), Coloring, Costas Array, Covering Array, Nonogram, Cril, Crossword, Dubois, Golomb Ruler, ii, insertion, Open Shop (os-taillard), Knights, Latin Square, Schurr's lemma, Magic Square, Mug, Myciel, Orthogonal Latin Squares, Quasi Group, Queen Attacking, Queen Knights, Radar Surveillance (Radar), Register, RLFAP-scen11 (scen11), Social Golfers, Super-Jobshop, Travelling Salesman Problem (TSP).
\end{quote}
We include all instances from each series except those which are not solved by all the heuristics used within timeout.

\begin{figure}
\centering
\includegraphics[scale=0.55]{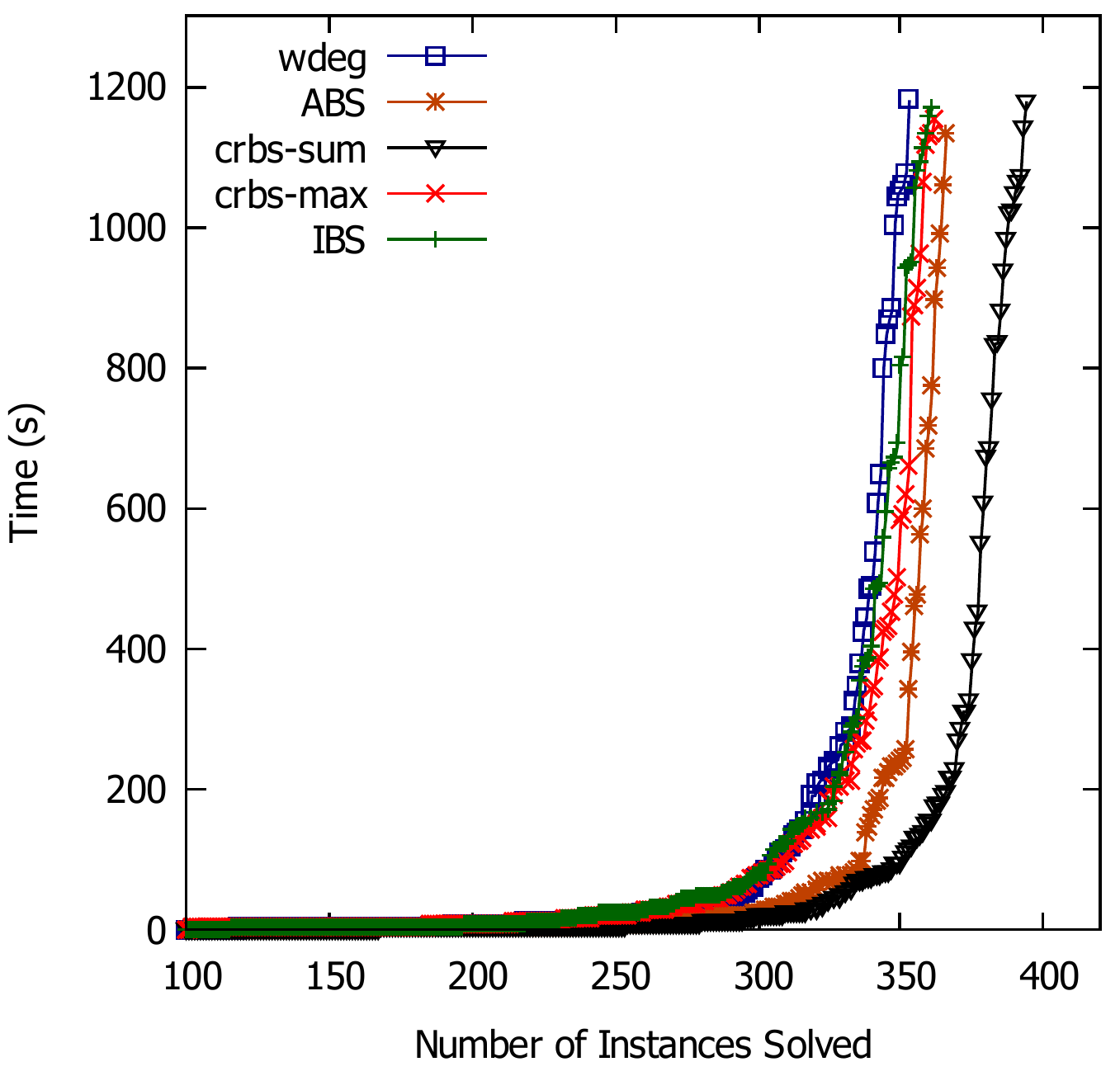}
\caption{Runtime distribution of all heuristics. \label{Fig_1}}
\end{figure}

\subsection{Comparing heuristics}

Figure \ref{Fig_1} shows a runtime distribution of the benchmark
instances solved using the different heuristics.
The y-axis is the CPU time (in seconds (s)) and the 
x-axis is the number of solved instances within the time limit. 
In the graph, instances which are too fast are ignored, namely, 
397 instances where the average time needed by
all heuristics is less than 1 second have not been plotted. 
Thus, there are 410 instances plotted in Figure \ref{Fig_1}.
Note that in this graph, the best performance is towards the lower right corner.

The best runtime distribution result is given by
the \textit{crbs-sum} heuristic which also solves the most instances.
In particular, with the time limit of 1200s,
\textit{crbs-sum} can solve 395/410 instances, which is better than \textit{dom/wdeg}, \textit{ABS}, \textit{IBS} and 
\textit{crbs-max} with respectively 354/410, 367/410, 362/410 and 363/410 instances.

Table \ref{exp:table1} gives the mean results of all five
heuristics on each series.
The row ``total ($n$)'' gives the average CPU times and number
of search nodes for all instances in a series, 
where $n$ is the number of instances.
The row ``solved by all ($n$)'' is the mean results on instances 
solved by all heuristics.
``$n$ TO'' denotes that the heuristic time-outs on $n$ instances. 
The bold numbers in Table \ref{exp:table1} highlight the best result 
for each series. 
Furthermore, for the Super-jobShop series, \textit{crbs-sum} has both
the smallest total time and smallest number of time-outs.
The last two rows, labelled as ``Total''
give the average results on all series 
for each heuristic with \textit{crbs-sum} giving the best overall results.

Table \ref{exp:table2} highlights how many series are solved faster
by a particular heuristic from the overall results
in Table \ref{exp:table1}.
The row ``Faster than \textit{dom/wdeg}'' (\textit{ABS} or \textit{IBS}) 
gives the number of series on which the heuristic is better than 
\textit{dom/wdeg}/\textit{ABS}/\textit{IBS} respectively.
The row ``Fastest (Second fastest)'' is the number of series on which 
the heuristic is the best (second best) respectively.
Overall, \textit{crbs-sum} solves more series.

The exact performance of the heuristics vary on different series.
\textit{crbs-sum} is the fastest on many series.
For example, on the dubois series, \textit{crbs-sum} solve 11 instances in 193 seconds, but \textit{dom/wdeg}, \textit{IBS} and \textit{ABS} time-out 
on some instances.
In total, \textit{crbs-sum}, \textit{crbs-max}, \textit{dom/wdeg}, \textit{IBS} and \textit{ABS} are the fastest on 10/30, 1/30, 6/30, 
9/30 
and 4/30 series respectively.
\textit{crbs-sum} is also competitive or better with the
other general purpose variable heuristics.
On 19 series, \textit{crbs-sum} is either the fastest or the second fastest heuristic.
Overall, \textit{crbs-sum} is faster than \textit{dom/wdeg}, \textit{ABS} and \textit{IBS} on 21, 20 and 19 series respectively.
For the Super-jobshop series, the mean times of \textit{crbs-sum} and \textit{dom/wdeg} are respectively 5.61s and 94.99s on the instances solved by the two heuristics.
The mean CPU time of \textit{crbs-sum} on all series is also less than that of other heuristics.

Between \textit{crbs-sum} and \textit{crbs-max}, our experiments show 
that the sum of correlations is more useful 
than the maximal correlation---\textit{crbs-sum} is faster 
than \textit{crbs-max} on many series.
On most series, the trend of mean times correlates with the trend on 
the number of nodes.
We observe that for the RLFAP-\textit{scen11} series, the number of
search nodes of \textit{crbs-sum} 
is less than that of \textit{dom/wdeg}, but \textit{crbs-sum} is slower than \textit{dom/wdeg}, thus, the cost of maintaining \textit{crbs-sum} may be more expensive than \textit{dom/wdeg}.
Possibly, our implementation could be optimized further.

\begin{table}
\small
\begin{center}
\tabcolsep=0.7mm
\vspace{0.4in}
 
\centering
\begin{tabular}{l|c|c|c|c|c}
\toprule [1 pt]
&\textit{dom/wdeg}&\textit{ABS}&\textit{IBS}&\textit{crbs-sum}&\textit{crbs-max}\\
\hline
Faster than \textit{dom/wdeg}&-&13&17&\textbf{21}&11\\
\hline
Faster than \textit{ABS}&16&-&\textbf{20}&\textbf{20}&14\\
\hline
Faster than \textit{IBS}&13&10&-&\textbf{19}&12\\
\hline
Faster than \textit{crbs-max}&19&16&18&\textbf{25}&-\\
\hline
Fastest&6&4&9&\textbf{10}&1\\
\hline
Second fastest&7&5&3&\textbf{9}&6\\
\toprule [1 pt]
\end{tabular}
\caption{Comparing heuristics.}  \label{exp:table2} 
 
\end{center}
\end{table}

\subsection{Choosing the crbs-sum parameter}
\label{sec:parameter}
\begin{figure}
\centering
\subfloat[TSP series]
{
\label{Fig_2a} 
\includegraphics[scale=0.29]{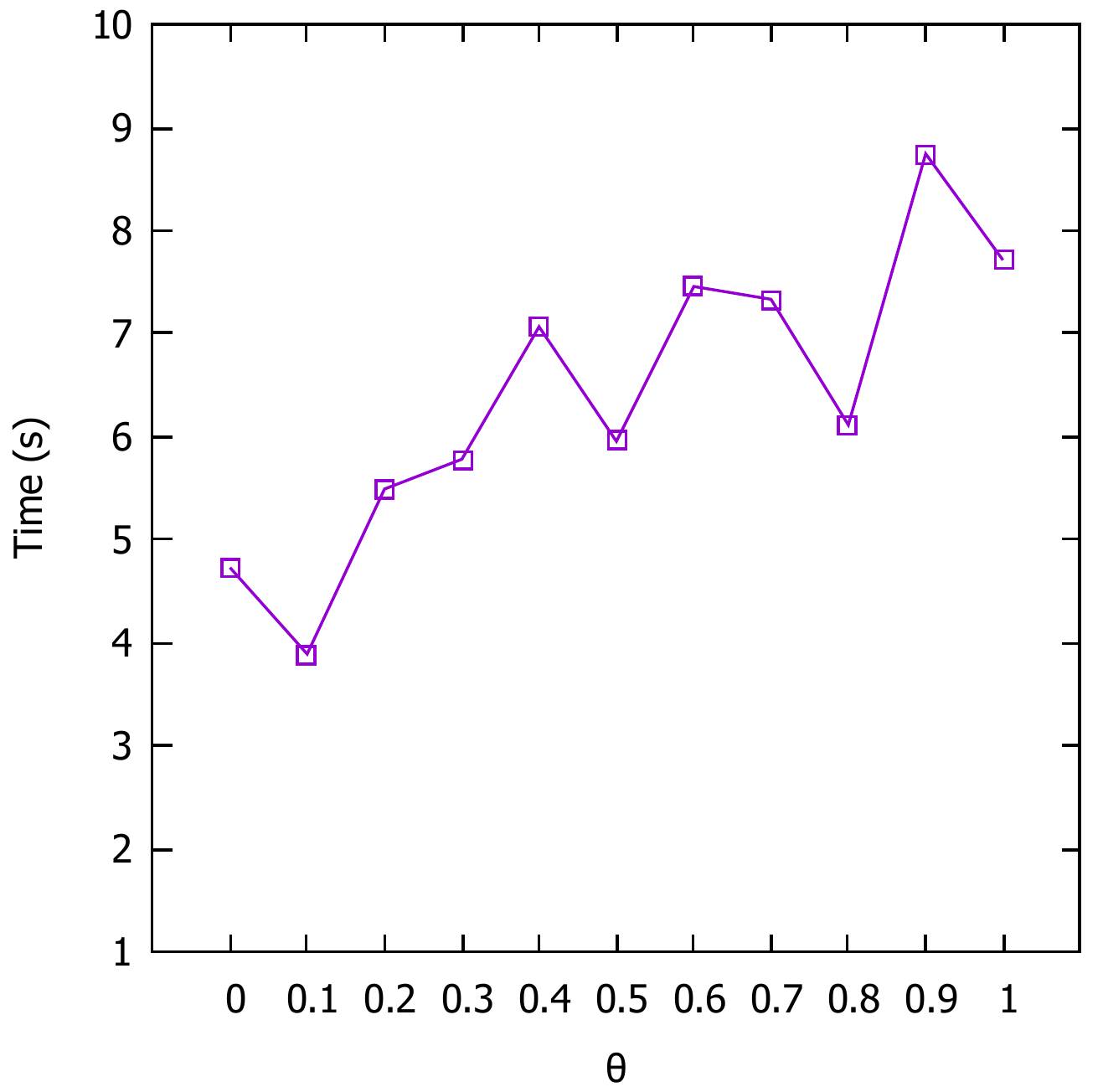}
}
\hspace{0in}
\subfloat[Quasi Group series]
{
\label{Fig_2b} 
\includegraphics[scale=0.29]{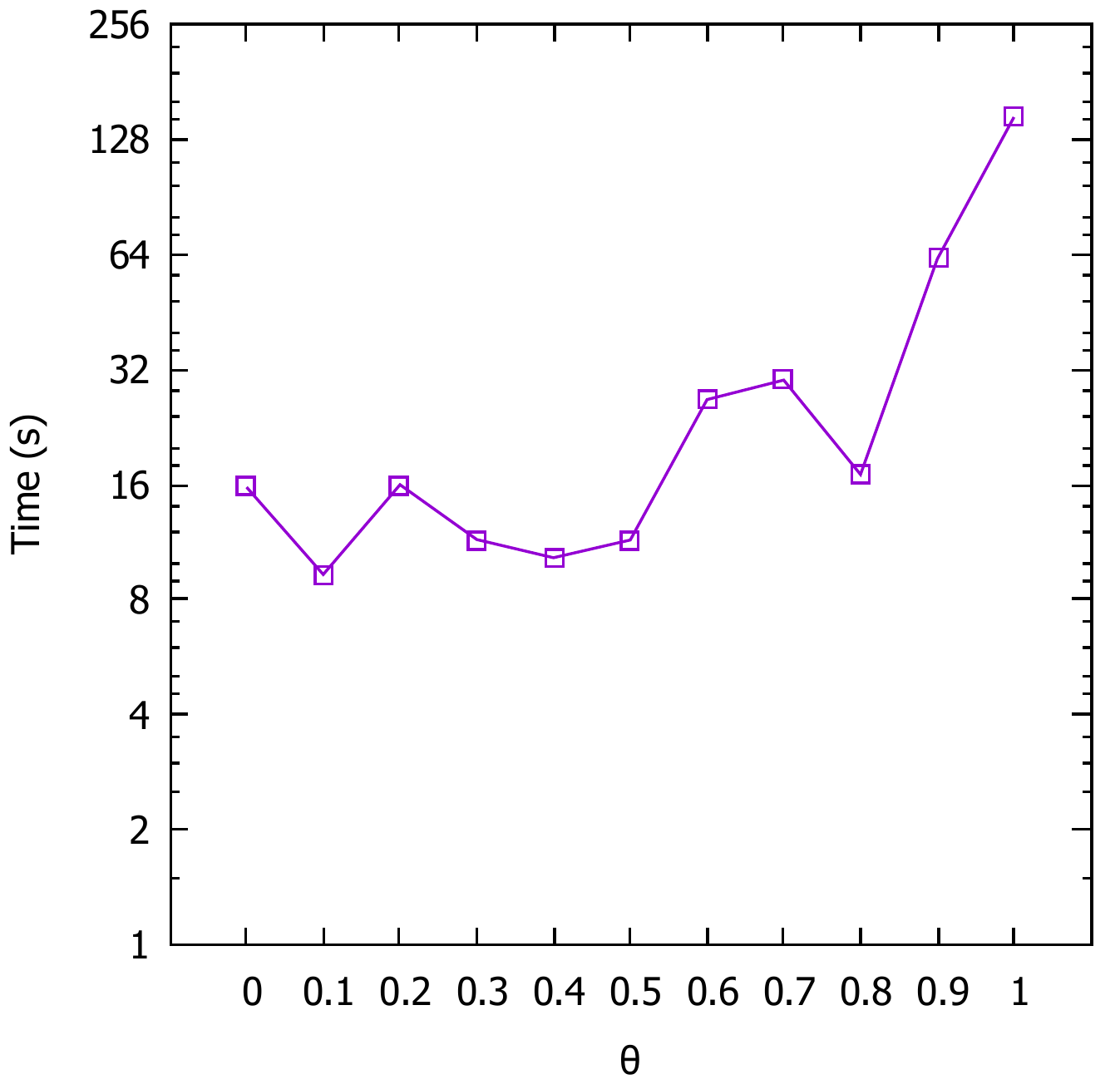}
}
\caption{Effect of $\theta$ on \textit{crbs-sum}.\label{Fig_2}}
\end{figure}

The parameter $\theta$ used in equation \ref{cor} affects the 
performance of the \textit{crbs-sum} heuristic.
In this section, we explore the effect of different choices of $\theta$
on two problem series.
Figure \ref{Fig_2a} gives the results on TSP series, 
where the y-axis is the mean solving times and the x-axis is the values of $\theta$. 
Correspondingly, Figure \ref{Fig_2b} gives
the results on the Quasi Group series.

Overall, we found that low values for the $\theta$ parameter generally
give better results than higher values.
For example, the mean times of solving TSP series is only 3.89s when $\theta=0.1$, which is 2 times faster than that of $\theta=0.9$. 
For extreme values of $\theta$,
when $\theta=0$, the mean time on the Quasi Group series is 9 times faster 
than that of $\theta=1$.
This suggests that the correlations between $x_i$ and past variables is more important for the \textit{crbs-sum} heuristic than correlations with future variables.
However, we also should not ignore the future variables, for example,
when $\theta=0.1$, the mean times on TSP and Quasi Group are faster than 
with $\theta=0$. 
  
\section{Conclusion}

In this paper, we propose a new idea, measuring correlations 
between variables, which leads to various correlation-based heuristics.
We measure and update the correlation matrix by using domain changes 
during constraint propagation.
We propose two correlation heuristics, 
\textit{crbs-sum} and \textit{crbs-max}, which employ different strategies
to estimate the potential of conflict for a variable 
based on the correlation matrix.
The experiments show that correlation heuristics are promising.
They are competitive with the state-of-the-art heuristics \textit{dom/wdeg}, \textit{ABS} and \textit{IBS} on a large set of benchmarks. 
Furthermore, the correlation heuristics can also be the fastest on 
many problem series.
In the future, we will explore more accurate or efficient methods to 
update the correlations between variables, and design 
improved correlation heuristics.

\section*{Acknowledgment}

This work has been supported by grant MOE2015-T2-1-117.

% trigger a \newpage just before the given reference
% number - used to balance the columns on the last page
% adjust value as needed - may need to be readjusted if
% the document is modified later
%\IEEEtriggeratref{8}
% The "triggered" command can be changed if desired:
%\IEEEtriggercmd{\enlargethispage{-5in}}

% references section

% can use a bibliography generated by BibTeX as a .bbl file
% BibTeX documentation can be easily obtained at:
% http://mirror.ctan.org/biblio/bibtex/contrib/doc/
% The IEEEtran BibTeX style support page is at:
% http://www.michaelshell.org/tex/ieeetran/bibtex/
\bibliographystyle{IEEEtran}
% argument is your BibTeX string definitions and bibliography database(s)
\bibliography{IEEEabrv,bibfile}
% <OR> manually copy in the resultant .bbl file
% set second argument of \begin to the number of references
% (used to reserve space for the reference number labels box)

% that's all folks
\end{document}